\documentclass[10pt,twocolumn,letterpaper]{article}

\usepackage{wacv}              % To produce the CAMERA-READY version
\usepackage[accsupp]{axessibility}
\definecolor{wacvblue}{rgb}{0.21,0.49,0.74}
\usepackage[pagebackref,breaklinks,colorlinks,allcolors=wacvblue]{hyperref}
\usepackage{tabularx}
\usepackage{arydshln}
\usepackage{multirow}
\usepackage{subcaption}
\usepackage{pifont}
\def\wacvPaperID{1596} % *** Enter the WACV Paper ID here
\def\confName{WACV}
\def\confYear{2026}

\title{GateFusion: Hierarchical Gated Cross-Modal Fusion for \\ Active Speaker Detection}

\author{Yu Wang \quad Juhyung Ha \quad Frangil M. Ramirez \quad Yuchen Wang \quad David J. Crandall\\
Indiana University\\
Bloomington, Indiana, USA\\
{\tt\small \{yw173, juhha, fraramir, wang617, djcran\}@iu.edu}
}

\begin{document}
\maketitle
\begin{abstract}
% Active Speaker Detection (ASD) seeks to determine who is speaking at each moment in a video.
Active Speaker Detection (ASD) aims to identify who is currently speaking in each frame of a video. 
% Active Speaker Detection (ASD) is a multimodal task that detects the speakers by correlating the audio stream with visual cues in a video.
%by modeling the complex interplay between audio and visual modalities. 
Most state-of-the-art approaches rely on late fusion to combine visual and audio features,
%only at highest semantic level, they 
but late fusion often fails to capture fine-grained cross-modal interactions, which can be critical for robust performance in unconstrained scenarios. In this paper, we introduce \textbf{GateFusion}, a novel architecture that combines strong pretrained unimodal encoders with a Hierarchical Gated Fusion Decoder (HiGate). HiGate enables progressive, multi-depth fusion by adaptively injecting contextual features from one modality into the other at multiple layers of the Transformer backbone, guided by learnable, bimodally-conditioned gates. To further strengthen multimodal learning, we propose two auxiliary objectives: Masked Alignment Loss (MAL) to align unimodal outputs with multimodal predictions, and Over-Positive Penalty (OPP) to suppress spurious video-only activations. GateFusion establishes new state-of-the-art results on several challenging ASD benchmarks, achieving 77.8\% mAP (+9.4\%), 86.1\% mAP (+2.9\%), and 96.1\% mAP (+0.5\%) on Ego4D-ASD, UniTalk, and WASD benchmarks, respectively, and delivering competitive performance on AVA-ActiveSpeaker. Out-of-domain experiments demonstrate the generalization of our model, while comprehensive ablations show the complementary benefits of each component.
\end{abstract}
    
\section{Introduction}
\label{sec:intro}

% Active Speaker Detection (ASD) is a multimodal task that detects the speakers by correlating the audio stream with visual cues in a video.
Active Speaker Detection (ASD) aims to detect who is speaking at each moment in a video.
% Active speaker detection (ASD) aims to detect speakers in each frame of a video. 
This is an inherently multimodal problem: both visual cues (e.g., whose mouth is moving, who is looking at whom) and audio (e.g., distinctive features of the speaker's voice) cues provide evidence for who is speaking~\cite{wang2024loconet, liao2023light, tao2021someone, alcazar2021maas, xiong2022looklistenmultimodalcorrelationlearning, lin2023quavf}, but individual modalities are often noisy or uninformative (e.g., the active speaker is not visible, multiple people have similar voices, etc.). Effectively modeling the interaction between these modalities remains a central challenge.

Existing approaches to audio-visual active speaker detection typically employ strong unimodal encoders and late fusion to combine features from different modalities. In such a framework, independent encoders process the audio and visual inputs, and their final representations are combined through a fusion module (such as summation or concatenation) before classification~\cite{liao2023light, tao2021someone, wang2024loconet, huh2025advancing, datta2022asd}. Although this strategy takes advantage of powerful unimodal backbones, it can fail to capture nuanced and hierarchical dependencies  between speech patterns and facial movements. As a result, these methods struggle in unconstrained or noisy environments where one modality may be uninformative or ambiguous.
%due to occlusions, background noise, or motion blur.

To address this gap, we propose a novel architecture that enables \textit{hierarchical cross-modal fusion} through a gating mechanism. Our key insight is that the final representation of one modality can be progressively enriched by integrating hidden features from the other modality across multiple encoder layers. Specifically, we introduce a decoder that performs coarse-to-fine fusion: at selected layers, the final output feature from the primary modality is injected with residual signals from the context modality. These injections are modulated by learnable gates, computed by the interaction of the primary modality feature with the hidden feature of the context modality, conceptually related to the gating idea in prior literature~\cite{jeong2025learning, qiu2025gated} but differing in both formulation and role. Our design is fully symmetric, allowing audio and visual modalities to each serve as either primary or context, thus supporting flexible bidirectional information flow. This design enables the model to selectively and adaptively incorporate multimodal information at various semantic levels.

Beyond the architecture, we introduce two auxiliary losses to further enhance the effectiveness and robustness of our model. First, we propose a Masked Alignment Loss (MAL) that encourages the unimodal predictions to align with the multimodal prediction  during periods of active speech by the target speaker. This  helps each modality remain consistent with the joint output, thereby improving generalization, especially in scenarios where one modality is ambiguous
(e.g., due to noise or occlusion). Second, we introduce the Over-Positive Penalty (OPP), which penalizes the model for excessive positive predictions from the visual stream alone. This regularization reduces false positives in visually complex or degraded scenes.

Experiments demonstrate that our model achieves state-of-the-art or competitive performance on both widely-used and emerging ASD benchmarks. Specifically, we surpass prior state-of-the-art results by 9.4\% on Ego4D-ASD~\cite{grauman2022ego4d} and 2.9\% on UniTalk~\cite{dataset_unitalk}, and attain the highest mAP of 96.1\% on WASD~\cite{dataset_wasd}. On  AVA-ActiveSpeaker~\cite{roth2020ava}, we achieve a strong 95.0\% mAP, the best result under the single-candidate setting.

In summary, our contributions are: (1) introducing a novel Hierarchical Gated Fusion Decoder (HiGate) that enables progressive, layer-wise cross-modal fusion through learnable gates; (2) proposing two auxiliary objectives, MAL for unimodal-multimodal alignment, and OPP to suppress visual-only false positives; and (3)  presenting a novel model, GateFusion, that establishes new state-of-the-art performance across both established and emerging ASD benchmarks.

\section{Related Work}
\label{sec:related_work}
\subsection{Active Speaker Detection (ASD)}

Recent progress on ASD has been fueled by large-scale multimodal datasets~\cite{roth2020ava, grauman2022ego4d, dataset_wasd, dataset_unitalk} and increasingly powerful unimodal models. Most existing approaches, including TalkNet~\cite{tao2021someone}, LightASD~\cite{liao2023light}, and ASD-Transformer~\cite{datta2022asd}, are based on separate encoders for audio and video streams with late fusion for classification. Although these designs are effective under controlled conditions, they cannot fully capture fine-grained interactions between modalities.

Models like ASDnB~\cite{roxo2024asdnb} and LASER~\cite{nguyen2025laser} incorporate additional cues such as body motion or lip landmarks to improve robustness. Other work~\cite{zhang2021unicon, wang2024loconet, min2022learning, alcazar2021maas, li2025egonet} has explored using multi-speaker contextual information, for instance through self-attention or graph-based aggregation across frames and speakers, as in LoCoNet~\cite{wang2024loconet} and SPELL~\cite{min2022learning}. 
However, since these methods  process multiple candidate face tracks simultaneously,  their computational cost is higher and they are   susceptible to degradation from low-quality faces in the same frame.
Other approaches focus on  challenging scenarios: EgoASD~\cite{huh2025advancing} applies data augmentation to encourage invariance to low-quality visual input, while rASD~\cite{vasireddy2024robust} employs explicit audio separation  to handle ambiguous or noisy acoustic signals.

In contrast, our proposed GateFusion model builds on robust pretrained encoders for both audio and visual modalities, but crucially, introduces hierarchical, gated cross-modal fusion across multiple representation depths. By aggregating and selectively injecting hidden features from low-level to high-level layers with adaptive gates, GateFusion can model fine-grained, context-aware interaction between modalities, which is critical for handling difficult or ambiguous inputs. Furthermore, rather than enforcing strong contrastive alignment as in TalkNCE~\cite{jung2024talknce}, GateFusion employs MAL to encourage unimodal-multimodal consistency and OPP to penalize overconfident visual predictions, promoting more calibrated and reliable outputs.

\subsection{Cross-Modality Fusion}
%Cross-modal fusion is a cornerstone of modern multi-modal learning, particularly in audio-visual tasks. 
Many recent papers~\cite{cheng2020look, tao2021someone, liao2023light, awan2024attend, wortwein2024smurf} use late fusion, where audio and visual features are extracted independently and merged at the final stage using  operations such as averaging or fully connected layers, as seen in Attend-Fusion~\cite{awan2024attend}. While these designs are straightforward and leverage powerful unimodal encoders, they often fail to fully exploit the rich interactions available between modalities.

To address these limitations, several studies explore explicit multimodal fusion at early, mid, or multi-layer stages. Owens and Efros~\cite{owens2018audio} propose to fuse audio and visual features at an early stage, while AV-SepFormer~\cite{lin2023av} performs multi-layer cross-modal fusion with a dedicated Transformer, combining fine- and coarse-grained modeling through intra-, inter-, and cross-modal blocks. Xu et al.~\cite{xu2024rethink} propose a messenger-guided mid-fusion Transformer, where modality-specific encoders exchange information via compact messenger tokens. CATNet~\cite{wang2024catnet} and MLCA-AVSR~\cite{wang2024mlca} both adopt multi-stage fusion, with different fusion block designs. DeepAVFusion~\cite{mo2024unveiling} employs an auxiliary fusion stream that interacts with audio and visual tokens from the earliest to the final layer.

Recent work such as AVIGATE~\cite{jeong2025learning} further explores gated cross-modal integration, combining video representations with audio cues using a gating mechanism for video-text retrieval. While AVIGATE inspired aspects of our approach, our proposed HiGate is distinct in several key ways.  HiGate treats audio and visual modalities as equally important, enabling symmetric and bidirectional cross-modal interaction. Moreover, rather than simply using a video query to attend to fixed audio features at each layer, HiGate injects hidden states from one modality into the final feature from another at selected hierarchical depths, modulated by learnable gates.
\begin{figure*}[h]
    \centering
    \includegraphics[width=0.7\linewidth]{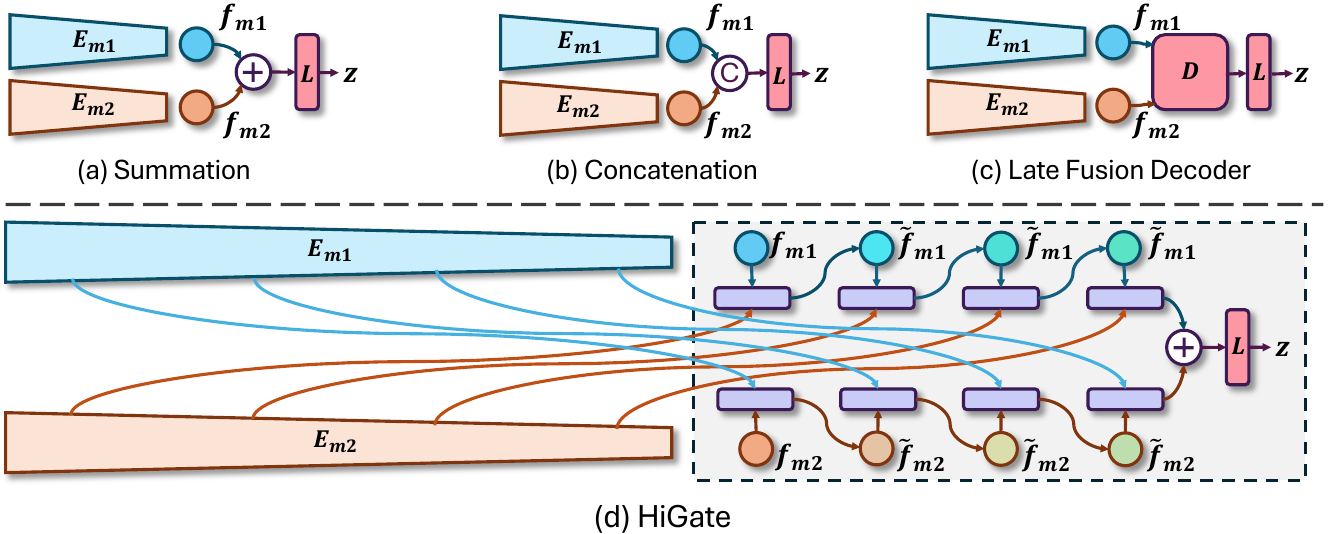}
    \caption{Overview of the proposed GateFusion architecture featuring HiGate. 
    Subfigures (a)-(c) depict typical late fusion strategies, where audio-visual features are extracted independently and fused only at the final stage: (a) \textit{Summation Fusion}, (b) \textit{Concatenation Fusion}, and (c) \textit{Late Fusion Decoder} (e.g., cross-attention) after unimodal encoding. In contrast, our method (d) \textit{HiGate} performs hierarchical cross-modal fusion by progressively injecting contextual signals from one modality into the other across multiple encoder layers. The degree of fusion is adaptively controlled by learnable gates, enabling fine-grained and robust audio-visual integration. For clarity, auxiliary unimodal classifiers used for computing the auxiliary losses are omitted from the illustration.}
    \label{fig:overall}
% \vspace{-5pt}
\end{figure*}
\section{Methodology}
We first describe the GateFusion architecture, focusing on the pretrained Transformer-based unimodal encoders. We then present  HiGate and its cross-modal fusion mechanism in detail. Finally, we introduce two auxiliary objectives, MAL and OPP, as well as  the final training objective.

\subsection{Multimodal Transformer}

GateFusion  performs frame-level active speaker estimation by jointly leveraging visual and auditory cues. We assume we are given a sequence of cropped facial video frames $I_v \in \mathbb{R}^{T_v \times H \times W}$ and the corresponding log-Mel spectrogram $I_a \in \mathbb{R}^{T_a \times N_{mels}}$ extracted from the audio. Each modality is first processed independently by a dedicated encoder. These encoders extract modality-specific features, which are then fused by the decoder and further processed by a linear classifier for final prediction.

For visual input we use  AV-HuBERT~\cite{shi2022learning}, which integrates a ResNet backbone~\cite{he2016deep} with a Transformer architecture for capturing complex spatiotemporal dependencies. For  audio, we use the Whisper encoder~\cite{radford2023robust}, which consists of an initial stack of convolutional layers to process spectral representations, followed by Transformer blocks to capture long-range temporal patterns. Both encoders operate without any discrete tokenization, preserving fine-grained modality information throughout the pipeline.

The output of each encoder is further processed by a projection head, implemented as a linear layer, to align its representation with the decoder. Since we use the short-time Fourier transform (STFT) for the audio stream, the temporal resolution of the audio features ($T_a$) is typically higher than that of the corresponding video frames ($T_v$). This means we need accurate temporal alignment between modalities to ensure meaningful fusion. 

Let $E_a$ and $E_v$ denote the audio and video encoders, each composed of $L$ Transformer blocks with $E^l$ representing the $l$-th block. Given the initial input $h^0$ (i.e., $I_a$ for audio or $I_v$ for video), the hidden states are computed as:
\begin{equation}
h^l = E^l(h^{l-1}),\quad l = 1,\ldots, L
\end{equation}
with final outputs $f_a = \phi(h_a^L)$ and $f_v = \phi(h_v^L)$. Here, $\phi$ denotes a projection operation using a single linear layer, reducing the unimodal features to the decoder width $F$. Their shapes are $f_a \in \mathbb{R}^{B \times T_a \times F}$ and $f_v \in \mathbb{R}^{B \times T_v \times F}$, where $B$ is the batch size.

\subsection{Hierarchical Gated Fusion Decoder (HiGate)}

Next, we introduce HiGate and how we integrate it with Transformer-based encoders. 

Previous state-of-the-art methods for ASD usually employ late fusion, in which audio and visual features are extracted independently using separate encoders and fused only at the final stage before prediction. As summarized in Fig.~\ref{fig:overall}(a)-(c), prevailing fusion strategies can generally be categorized into three main types: summation, concatenation, and late fusion decoders.
These conventional approaches implicitly assume that the unimodal encoders, often pretrained on large-scale datasets,
consistently produce rich representations that can fully encode complementary cues across modalities. However, as information propagates through deep Transformer layers, fine-grained information is often lost, with only abstract high-level semantics remaining in the final representation. This means that important information may be lost before the cross-modality fusion occurs. 
Moreover, the output of each unimodal encoder can be significantly affected by noisy, occluded, or low-resolution inputs.

To address these limitations, our HiGate decoder (Fig.~\ref{fig:overall}(d)) hierarchically enriches the representations from each encoder by integrating selected hidden states from the corresponding Transformer blocks. This approach allows for symmetric, hierarchical cross-modal fusion: either modality can serve as the primary or the context, providing a flexible and generalizable framework. Specifically, we select a set of hidden states $\{h_c^{l}\}$, where $l \in \{l_0, l_1, \ldots, l_N\}$ and $N \leq L$, from the context modality's Transformer blocks. The final output feature $f_p$ of the primary modality encoder is then sequentially refined through interaction with these context hidden states.

\begin{figure}[t]
    \centering
    \includegraphics[width=1\linewidth]{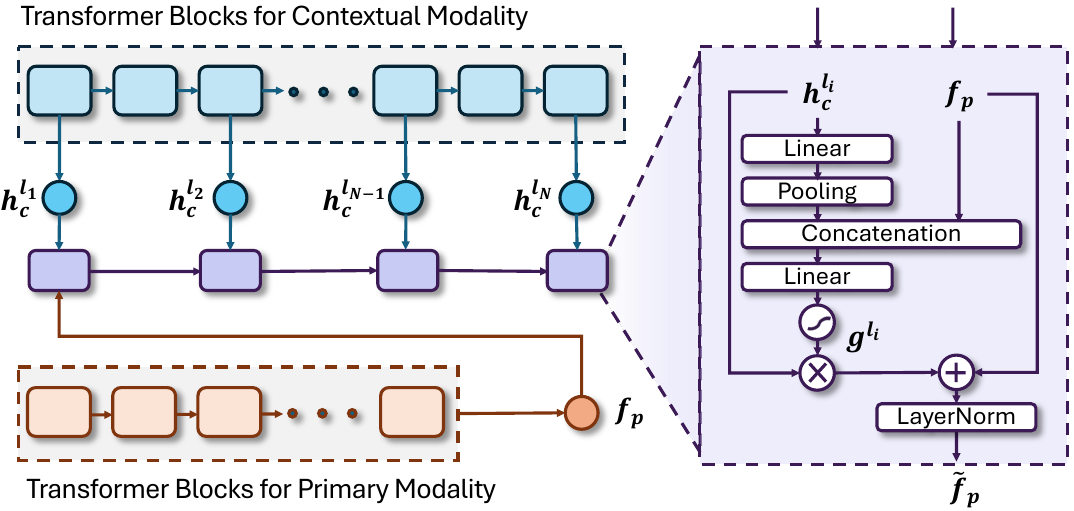}
    \caption{Illustration of the gating mechanism. The initial output $f_p$ from the primary modality is progressively integrated with hidden states $h_c^{l}$ from the context modality at multiple layers, with each fusion step modulated by a learnable gate. This process iteratively refines $f_p$ into the final enriched representation $\tilde{f}_p$.}
    
    \label{fig:gate}
\end{figure}

Formally, for each selected hidden state $h_c^{l}$ from the context modality and the current feature $\tilde{f}_p$ of the primary modality (with $\tilde{f}_p = f_p$ initially), we first project $h_c^{l}$ to match the feature dimension of $\tilde{f}_p$ and then align the temporal dimension,
\begin{equation}
    \tilde{h}_c^l = \mathcal{I}(\phi(h_c^{l}), T_{p}),
\end{equation}
where $\mathcal{I}(I, T)$ indicates temporal alignment of $I$ to the target size $T$ via average pooling.

Next, the aligned context feature $\tilde{h}_c^l$ is injected into $\tilde{f}_p$ under the control of a learnable gate. This gate determines the degree to which information is borrowed from the context hidden state, and is computed through the interaction between the features of the primary modality and those of the context modality,
\begin{equation}
    % g^l = \sigma(\mathrm{MLP}(\mathrm{Concat}(\tilde{f}_p, \tilde{h}_c^l))),
    g^l = \sigma(W_{g} [\tilde{f}_p; \tilde{h}_c^l]),
\end{equation}
where $\sigma$ denotes the sigmoid activation, $[\cdot ; \cdot]$ represents feature concatenation, and $W_{g}$ is a learnable weight matrix in a linear layer.  The fusion at each selected layer is then performed,
\begin{equation}
    \tilde{f}_p = \mathrm{LN}(\tilde{f}_p + g^l \odot \tilde{h}_c^l),
\end{equation}
where $\mathrm{LN}$ denotes Layer Normalization and $\odot$ denotes element-wise multiplication. Through this process, the primary modality's representation is progressively enhanced by gated information from the context modality at multiple semantic levels, which we hypothesize will lead to more robust and fine-grained audio-visual integration for ASD. The procedure is illustrated in Fig.~\ref{fig:gate}.

\subsection{Training Losses}
To further enhance the robustness and generalization of GateFusion, we introduce two auxiliary objectives: MAL and OPP. These losses are designed to reinforce multimodal alignment and improve unimodal reliability, particularly in challenging audio-visual scenarios.

In addition to the main multimodal prediction head, we attach a linear classifier to the final representations of each unimodal branch (audio and video), yielding per-frame logits for speaker activity. During preliminary experiments, we noticed that when unimodal classifiers are trained either with or without explicit supervision, their predictions can become misaligned with those of the multimodal output, particularly in challenging or ambiguous conditions. This misalignment suggests that the unimodal branches may not fully use cross-modal information available during training, reducing their reliability as fallback predictors when one modality is not informative.

\subsubsection{Masked Alignment Loss (MAL)}
To address this, we propose MAL, which enforces consistency between the unimodal  and  multimodal predictions, but only on frames where active speakers are present. This is motivated by the fact that  negative frames may contain voices from off-screen speakers or other background sources, leading to ambiguous supervision. By focusing alignment only on positive frames, we encourage unimodal branches to model the cross-modal dependencies required for accurate speaker activity detection, while avoiding overfitting to modality-specific artifacts.
Specifically, 
\begin{align}
    f_{av} &= \mathcal{I}(\tilde{f}_a, T_v) + \tilde{f}_v \\
    \mathbf{p}_{av} &= \mathrm{softmax}(\mathrm{stop\_grad}(\mathcal{C}_{av}(f_{av}))) \\
    \mathbf{p}_a &= \mathrm{softmax}(\mathcal{C}_{a}(\mathcal{I}(f_a, T_v))) \\
    \mathbf{p}_v &= \mathrm{softmax}(\mathcal{C}_{v}(f_v)),
\end{align}
where $\mathrm{stop\_grad}$ denotes the stop-gradient operation which prevents gradients from flowing into the multimodal branch during alignment and $\mathcal{C}_{av}$, $\mathcal{C}_{a}$ and $\mathcal{C}_{v}$ denote separate linear classifiers.

The set of positive frames is denoted as $S = \{i \mid y^{(i)} = 1\}$, where $y^{(i)} \in \{0, 1\}$ is the ground-truth speaker activity label for frame $i$. The MAL penalizes the KL divergence between the multimodal prediction and each unimodal output, averaged only over the positive frames,
\begin{align}
\mathrm{KL}_a^{(i)} &= \sum_{c=1}^{C} \mathbf{p}_{av}^{(i)}[c] \left( \log \mathbf{p}_{av}^{(i)}[c] - \log \mathbf{p}_a^{(i)}[c] \right) \\
\mathrm{KL}_v^{(i)} &= \sum_{c=1}^{C} \mathbf{p}_{av}^{(i)}[c] \left( \log \mathbf{p}_{av}^{(i)}[c] - \log \mathbf{p}_v^{(i)}[c] \right) \\
\mathcal{L}_{\text{MAL}} &= \frac{1}{2|S|} \left( \sum_{i \in S} \mathrm{KL}_a^{(i)} + \sum_{i \in S} \mathrm{KL}_v^{(i)} \right),
\end{align}
where $c$ indexes the class ($C = 2$ for binary classification).
The positive mask is important, as negative frames may correspond to visible faces without active speech or off-camera voices, making alignment less meaningful.

\subsubsection{Over-Positive Penalty (OPP)}
We  noticed that the video branch is prone to excessive false positives, particularly in egocentric data where visual ambiguity and motion blur are common. We thus introduce OPP, which penalizes the video classifier for predicting speaker activity in frames where the ground truth indicates silence,
\begin{align}
\mathcal{L}_{\text{OPP}} &= \frac{1}{M} \sum_{i=1}^M \mathbf{p}_{v}^{(i)}[1] \cdot (1 - y^{(i)}).
\end{align}
Here, $\mathbf{p}_{v}^{(i)}[1]$ is the predicted probability of positive label (active speech) by the video branch for frame $i$, and $M$ is the total number of frames. This loss encourages the video stream to maintain calibrated confidence and suppress spurious activations in visually ambiguous scenarios.

\subsubsection{Final Objective}
We hypothesize that together, MAL and OPP help to improve multimodal synergy and reliable unimodal fallback, and will lead to better ASD under challenging conditions.

Our primary classification objective is the cross-entropy loss over multimodal predictions without a stop-gradient operation,
\begin{equation}
\mathcal{L}_{\text{CLS}} = -\frac{1}{M} \sum_{i=1}^{M} \sum_{c=1}^{C} \hat{y}^{(i)}[c] \log \mathbf{p}_{av}^{(i)}[c],
\end{equation}
where $\hat{y}^{(i)} \in \{0,1\}^C$ is the one-hot encoding of $y^{(i)}$. The final training objective is,
\begin{equation}
    \mathcal{L} = \mathcal{L}_{\text{CLS}} + \lambda_{\text{MAL}} \mathcal{L}_{\text{MAL}} + \lambda_{\text{OPP}} \mathcal{L}_{\text{OPP}},
\end{equation}
where $\lambda_{\text{MAL}}$ and $\lambda_{\text{OPP}}$ are hyperparameters that control the influence of the auxiliary losses.

\section{Results}
We first detail our experimental setup, followed by  performance evaluations, generalization to out-of-domain datasets, and  an ablation study.

\subsection{Datasets}
We evaluate our method on four diverse large-scale datasets covering egocentric, cinematic, and unconstrained scenarios.

\vspace{6pt}\noindent
\textbf{Ego4D}~\cite{grauman2022ego4d} is a large-scale collection of egocentric videos. We use its audio-visual benchmark subset, denoted \textbf{Ego4D-ASD}, which contains 50 hours of video, presenting challenges like frequent camera motion.

\vspace{6pt}\noindent
\textbf{AVA-ActiveSpeaker}~\cite{roth2020ava} contains 38.5 hours of Hollywood movie clips featuring variations in lighting, occlusion, and background speech.

\vspace{6pt}\noindent
%\subsubsection{Wilder Active Speaker Detection (WASD) Dataset}
\textbf{Wilder Active Speaker Detection (WASD)}~\cite{dataset_wasd} comprises approximately 30 hours of YouTube videos, covering five challenging categories: Optimal, Speech Impairment, Face Occlusion, Human Voice Noise, and Surveillance.

\vspace{6pt}\noindent
%{UniTalk Dataset}
\textbf{UniTalk}~\cite{dataset_unitalk} is an emerging large-scale benchmark containing over 44.5 hours of in-the-wild YouTube videos. It targets four specific challenges: underrepresented languages, high ambient noise, crowded scenes, and difficult combinations of these factors.

\subsection{Experimental Setup}
Following standard preprocessing, we extracted face crops using the provided bounding boxes and resized them to $112 \times 112$ pixels. These were temporally stacked to form the visual input sequence. The corresponding audio was converted into log-Mel spectrograms using the official Whisper log-Mel extraction pipeline~\cite{radford2023robust}. We initialized the visual and audio encoders with pretrained weights from Noise-Augmented AV-HuBERT Large \cite{shi2022robust} and Whisper large-v3~\cite{radford2023robust}, respectively. To reduce computation while preserving important temporal and spatial cues, we only retained the first 12 Transformer layers of each encoder, based on our empirical findings that low-level representations capture discriminative features for ASD. We used a two-layer MLP classifier with GELU activation for multimodal prediction, and separate single-layer linear classifiers for the audio and video branches to compute the MAL and OPP losses.

We trained all models using the AdamW optimizer~\cite{loshchilov2017decoupled}, with a batch size of 1,500 frames and a total of 30,000 training steps. Learning rates were set to $5 \times 10^{-5}$ for the encoders and $1 \times 10^{-4}$ for the decoder, with a decay factor of 0.95 applied every 3,000 steps. All experiments were conducted on a single NVIDIA L40S GPU.

For our HiGate decoder, we selected four fusion points at encoder layers $\{1, 4, 7, 10\}$, with the index starting from 1. The impact of this selection strategy is analyzed in Sec.~\ref{sec:ablation_studies}. The decoder width was set to 1280 for all experiments. We empirically set the loss weights to $\lambda_{\text{MAL}}=0.01$ and $\lambda_{\text{OPP}}=0.1$ to balance the auxiliary supervision with the primary classification objective.

\subsection{Comparison with the State-of-the-art}
We evaluate GateFusion on four challenging ASD benchmarks and compare with state-of-the-art methods.  
\begin{table}[t]
\centering
\small
\begin{tabularx}{\columnwidth}{X l}
\toprule
\textbf{Model} & \textbf{mAP} \\
\midrule
TalkNet~\cite{tao2021someone}      & 51.7 \\
LightASD~\cite{liao2023light}     & 52.8$^\dagger$ \\
SPELL~\cite{min2022intel, min2022learning}        & 60.7 \\
STHG~\cite{min2023sthg}         & 63.7 \\
LoCoNet~\cite{wang2024loconet}      & \underline{68.4} \\
\midrule
GateFusion & \textbf{77.8} \\
\bottomrule
\end{tabularx}
\caption{Comparison of mAP scores on the Ego4D-ASD benchmark~\cite{grauman2022ego4d}. Scores marked with $^\dagger$ were obtained by us using the official implementation.}
\label{table:ego4d}
\vspace{-5pt}
\end{table}

\vspace{6pt}\noindent
\textbf{Ego4D-ASD.} Table~\ref{table:ego4d} shows results on the Ego4D-ASD benchmark, which features egocentric and visually challenging scenarios from highly dynamic scenes. GateFusion achieves a new state-of-the-art mAP of 77.8\%, outperforming the previous best method, LoCoNet~\cite{wang2024loconet}, by over 9 points. Earlier baselines such as TalkNet~\cite{tao2021someone} and LightASD~\cite{liao2023light} show inferior performance, highlighting the robustness of our hierarchical fusion strategy under dynamic egocentric conditions. Since LightASD~\cite{liao2023light} did not originally report results on Ego4D-ASD, we reproduced their score using the authors' public implementation.

\begin{table}[t]
  \centering
  \footnotesize
  \begin{tabular*}{\columnwidth}{@{\extracolsep{\fill}}lccccc}
    \toprule
    \textbf{Model} & \textbf{Overall} & \textbf{Language} & \textbf{Crowded} & \textbf{Noise} & \textbf{Hard} \\
    \midrule
    ASDNet~\cite{kopuklu2021design}        & 20.6 & 30.8 & 17.5 & 14.8 & 20.3 \\
    ASC~\cite{alcazar2020active}           & 61.4 & 74.7 & 62.9 & 53.4 & 57.3 \\
    TalkNet~\cite{tao2021someone}          & 75.7 & 80.1 & 77.6 & 67.1 & 70.3 \\
    LoCoNet~\cite{wang2024loconet}         & 82.2 & 85.8 & 84.6 & 80.0 & 76.2 \\
    TalkNCE~\cite{jung2024talknce}         & \underline{83.2} & \underline{86.7} & \underline{84.9} & \underline{84.1} & \underline{77.9} \\
    \midrule
    GateFusion                             & \textbf{86.1} & \textbf{89.6} & \textbf{87.8} & \textbf{88.7} & \textbf{79.9} \\
    \bottomrule
  \end{tabular*}
  \caption{Comparison of mAP scores on UniTalk~\cite{dataset_unitalk}. \textbf{Overall} denotes the average over the full test set. \textbf{Language} includes scenes where the primary spoken language is non-English, \textbf{Noise} scenes have high ambient noise, \textbf{Crowded} scenes have multiple visible speakers, and \textbf{Hard} scenes have multiple challenging conditions.}
  %\vspace{-0.5cm}
  \label{tab:unitalk}
\end{table}

\vspace{6pt}\noindent
\textbf{UniTalk.} Table~\ref{tab:unitalk} summarizes performance on this large-scale, multi-condition benchmark that includes subsets for under-represented spoken language, crowded scenes, ambient noise level, and combinations of multiple challenging factors. GateFusion achieves an overall mAP of 86.1\%, outperforming all prior approaches across every category, with notable gains under Noise and Hard conditions, demonstrating robustness in degraded environments.

\vspace{6pt}\noindent
\textbf{WASD.} Table~\ref{table:wasd} reports mAP scores on the WASD benchmark, which evaluates resilience across real-world conditions including speech impairments, face occlusions, ambient human voices, and surveillance camera settings. GateFusion achieves the highest overall mAP of 96.1\% and sets new state-of-the-art scores in nearly all subcategories, working especially well under surveillance and voice-noise conditions.

\begin{table}[t]
    \centering
    \small
    \renewcommand{\arraystretch}{1.05}
    \begin{tabular}{l*{6}{c}}
        \toprule
        \multirow{2}{*}{\textbf{{Model}}} &
        \multicolumn{2}{c}{{\textbf{Easy}}} & 
        \multicolumn{3}{c}{{\textbf{Hard}}} & \multirow{2}{*}{\textbf{All}} \\
        \cmidrule(r){2-3} \cmidrule(r){4-6}
        & \textbf{OC} & \textbf{SI} & \textbf{FO} & \textbf{HVN} & \textbf{SS} & \\
        \hline
        ASC~\cite{alcazar2020active} & 
        91.2 & 92.3 & 87.1 & 66.8 & 72.2 & 85.7\\
        MAAS~\cite{alcazar2021maas} & 
        90.7 & 92.6 & 87.0 & 67.0 & 76.5 & 86.4\\
        ASDNet~\cite{kopuklu2021design} & 
        96.5 & 97.4 & 92.1 & 77.4 & 77.8 & 92.0\\
        TalkNet~\cite{tao2021someone} & 
        95.8 & 97.5 & 93.1 & 81.4 & 77.5 & 92.3\\
        TS-TalkNet~\cite{ts_talknet} & 
        96.8 & 97.9 & 94.4 & 84.0 & 79.3 & 93.1\\
        LightASD~\cite{liao2023light} & 
        97.8 & 98.3 & 95.4 & 84.7 & 77.9 & 93.7\\
        BIAS~\cite{method_bias} & 
        97.8 & 98.4 & \underline{95.9} & 85.6 & 82.5 & 94.5 \\
        ASDnB~\cite{roxo2024asdnb} & %
        \underline{98.7} & \underline{98.9} & \textbf{97.2} & \underline{89.5} & \textbf{82.7} & \underline{95.6}\\
        \midrule
        GateFusion &
        \textbf{98.9} & \textbf{99.2} & \textbf{97.2} & \textbf{92.1} & \underline{82.6} & \textbf{96.1}\\
        % Ours-full &
        % 99.3 & 99.5 & 98.1 & 94.3 & 86.3 & 97.1 \\
        % Ours-full &
        \bottomrule
    \end{tabular}
     \caption{Comparison of mAP scores on WASD~\cite{dataset_wasd}. \textbf{OC} refers to Optimal Conditions, \textbf{SI} to Speech Impairment, \textbf{FO} to Face Occlusion, \textbf{HVN} to Human Voice Noise, \textbf{SS} to Surveillance Settings, and \textbf{All} to the average performance across the entire test set.}
     %\vspace{-0.5cm}
    \label{table:wasd}
\end{table}

\vspace{6pt}\noindent
\begin{table}[t]
\centering
\small
\begin{tabularx}{\columnwidth}{X c  c}
\toprule
\textbf{Method} & \textbf{Candidates} & \textbf{mAP} \\
\midrule
Sync-TalkNet~\cite{wuerkaixi2022rethinking}     & Single   & 88.8 \\
TalkNet~\cite{tao2021someone}     & Single   & 92.3 \\
ASD-Transformer~\cite{datta2022asd} & Single   & 93.0 \\
LightASD~\cite{liao2023light}   & Single   & 94.1 \\
LR-ASD~\cite{liao2025lr}   & Single   & 94.5 \\
\midrule
MAAS~\cite{alcazar2021maas}     & Multiple   & 88.8 \\
EASEE-50~\cite{alcazar2022end}     & Multiple   & 94.1 \\

SPELL~\cite{min2022intel,min2022learning} & Multiple & 94.2 \\
STHG~\cite{min2023sthg}        & Multiple & 94.9 \\
LoCoNet~\cite{wang2024loconet}  & Multiple & \textbf{95.2} \\
\midrule
GateFusion                          & Single   & \underline{95.0} \\
% GateFusion (Large)                 & Single   & \textbf{95.9} $ \ $ \\
\bottomrule
\end{tabularx}
\caption{Comparison of mAP on the AVA-ActiveSpeaker benchmark~\cite{roth2020ava}. \textbf{Candidates} denotes whether evaluation is performed under the single-candidate or multi-candidate setting.}
\label{table:ava}
% \vspace{-10pt}
\end{table}

\vspace{6pt}\noindent
\textbf{AVA-ActiveSpeaker.} Table~\ref{table:ava} presents results on the widely adopted AVA-ActiveSpeaker benchmark. The ``Candidates'' column indicates whether each model is evaluated under the single-candidate or multi-candidate setting. The multi-candidate setting provides additional scene context by jointly considering multiple faces. In the single-candidate setting, GateFusion attains an mAP of 95.0\%, surpassing the previous state-of-the-art method, LR-ASD~\cite{liao2025lr}, by 0.5\%. Although LoCoNet~\cite{wang2024loconet} slightly outperforms GateFusion under the multi-candidate configuration, this gain is attributable to the extra contextual information and the unique nature of the AVA-ActiveSpeaker dataset, which features stable, well-separated multi-speaker interactions. Such conditions favor context-aggregation models like LoCoNet~\cite{wang2024loconet}. However, most other benchmarks lack this level of consistent multi-person context. %especially for the egocentric data, thereby constraining the generalizability of multi-candidate approaches.

Overall, GateFusion consistently delivers state-of-the-art or highly competitive performance across all benchmarks, demonstrating robustness in both controlled environments and unconstrained conditions. These results highlight the model’s strong generalization across diverse audio-visual contexts without relying on explicit multi-person cues. 

In Supplementary Sec.\ A, we detail the initialization strategies of all competing methods and further compare against two representative ASD models that share the same pretrained encoders as ours. These results suggest that the performance gains are not solely attributable to encoder pretraining, but primarily arise from the proposed fusion design. 

\subsection{Comparison of Generalization Performance}
To assess the generalization ability of ASD models, we evaluate each approach in both in-domain (internal) and out-of-domain (external) settings, as summarized in Tables~\ref{tab:unitalk2ava} and~\ref{tab:ava2wasd}. Internal evaluation refers to models trained and tested on the same dataset, while external evaluation measures performance when the model is applied to a distinct, unseen dataset.

\begin{table}[t]
    \centering
    \begin{subtable}[h]{\columnwidth}
        \centering
      \begin{tabular}{lcc}
        \toprule
        \textbf{{Model}} & \textbf{UniTalk (Internal)} & \textbf{AVA (External)} \\
        \midrule
        TalkNet~\cite{tao2021someone}    & 75.7 & 78.4 \\
        LoCoNet~\cite{wang2024loconet}   & 82.2 & 84.4\\
        TalkNCE~\cite{jung2024talknce}   & \underline{83.2} & \underline{88.0}\\
        \midrule
        GateFusion   & \textbf{86.1} & \textbf{90.7}\\
        \bottomrule
      \end{tabular}
       \caption{UniTalk~\cite{dataset_unitalk} $\rightarrow$ AVA-ActiveSpeaker~\cite{roth2020ava}}
       \label{tab:unitalk2ava}
    \end{subtable}
    \hfill
    \vspace{6pt}
    \begin{subtable}[h]{\columnwidth}
        \centering
      \begin{tabular}{lcc}
        \toprule
        \textbf{Model} & \textbf{AVA (Internal)} & \textbf{WASD (External)} \\
        \midrule
        ASC~\cite{alcazar2020active}        & 83.6 & 74.6 \\
        MAAS~\cite{alcazar2021maas}         & 82.0 & 70.7 \\
        ASDNet~\cite{kopuklu2021design}     & 91.1 & 79.2 \\
        TalkNet~\cite{tao2021someone}       & 91.8 & 85.0 \\
        TS-TalkNet~\cite{ts_talknet}        & 92.7 & 85.7 \\
        LightASD~\cite{liao2023light}      & \underline{93.4} & \underline{86.2} \\
        \midrule
        GateFusion                          & \textbf{95.0} & \textbf{88.8} \\
        \bottomrule
      \end{tabular}        
        \caption{ AVA-ActiveSpeaker~\cite{roth2020ava} $\rightarrow$ WASD~\cite{dataset_wasd}}
        \label{tab:ava2wasd}
     \end{subtable}
     \caption{Generalization of model performance on out-of-domain datasets. \textbf{Internal} refers to performance when the model is both trained and tested on the same dataset, while \textbf{External} measures performance when the model is tested on an out-of-domain dataset to assess generalization.}
     %\vspace{-0.5cm}
     \label{tab:generalization}
\end{table}

Table~\ref{tab:unitalk2ava} compares performance when models are trained on UniTalk and tested either in-domain (UniTalk) or cross-domain (AVA-ActiveSpeaker). GateFusion achieves the highest mAP in both settings, with 86.1\% on UniTalk and an especially strong 90.7\% on AVA-ActiveSpeaker, outperforming previous state-of-the-art methods including TalkNet~\cite{tao2021someone}, LoCoNet~\cite{wang2024loconet}, and TalkNCE~\cite{jung2024talknce}. The large improvement in external evaluation demonstrates GateFusion's ability to transfer knowledge and maintain performance across diverse domains.

Table~\ref{tab:ava2wasd} presents a complementary evaluation, in which models are trained on AVA-ActiveSpeaker and tested on WASD. Again, GateFusion consistently outperforms all competing baselines, achieving 95.0\% in-domain (AVA-ActiveSpeaker) and 88.8\% on the out-of-domain WASD benchmark. In both internal and external scenarios, GateFusion not only surpasses architectures such as ASC~\cite{alcazar2020active}, MAAS~\cite{alcazar2021maas}, and ASDNet~\cite{kopuklu2021design}, but also outperforms recent strong baselines including LightASD~\cite{liao2023light}.

Collectively, these results demonstrate that GateFusion delivers state-of-the-art generalization across domains, demonstrating robustness to distribution shifts and supporting its practical utility in unconstrained environments.
\begin{figure}[t]
  \centering
  \includegraphics[width=1\linewidth]{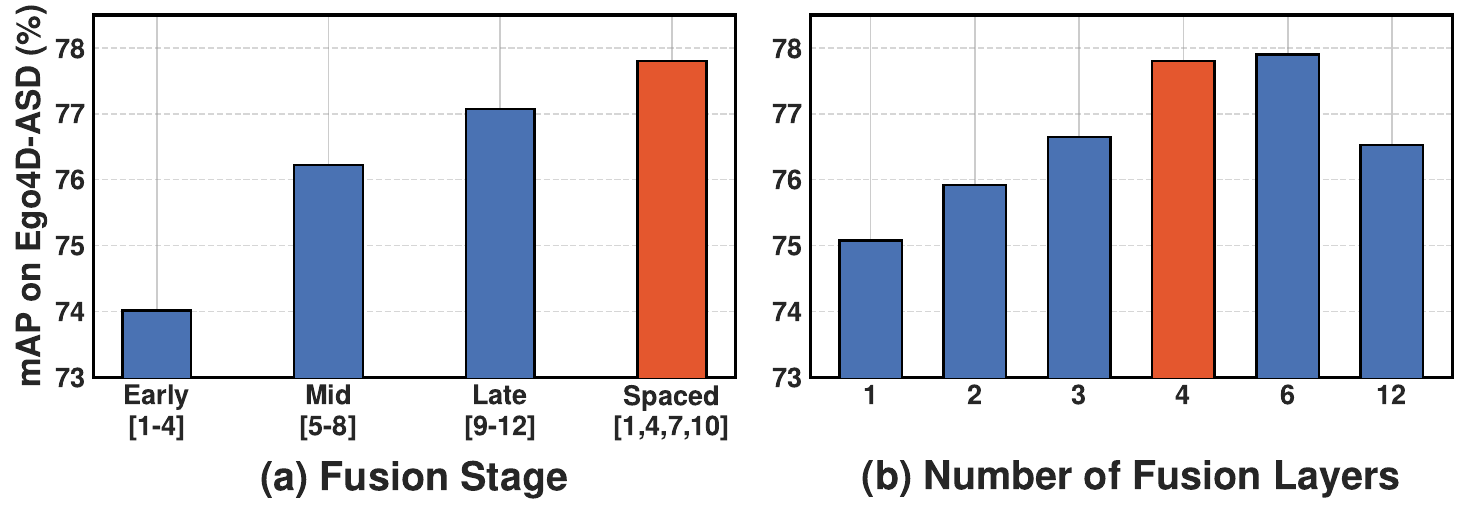}
  \caption{Ablations on (a) fusion stage and (b) number of fusion layers on Ego4D-ASD, showing the impact of fusion policies. Orange bars mark our chosen configuration (layers 1, 4, 7, 10).}
  \label{fig:ablation_fusion}
\end{figure}

\subsection{Ablation Studies}
\label{sec:ablation_studies}
\textbf{Fusion-Policy Analysis.} Fig.~\ref{fig:ablation_fusion} summarizes a sensitivity analysis on (a) fusion stage and (b) number of fusion layers. Late-stage fusion yields stronger performance than early or mid fusion alone, indicating that deeper features provide stronger cues. A spaced policy that combines shallow and deep fusion points provides the best mAP, demonstrating that early and mid layers provide complementary information. Furthermore, we observe that increasing the number of fusion layers improves performance up to six layers; however, utilizing four spaced fusion points yields an optimal trade-off between accuracy and efficiency. More details are provided in Supplementary Sec.\ B.

\begin{table}[t]
% \vspace{-1em}
\centering
\small
% \vspace{-0.5em}
\setlength{\tabcolsep}{6pt}
\begin{tabular}{ccc|cccc}
\toprule
\multicolumn{3}{c|}{\textbf{Configuration}} & \textbf{Ego4D-ASD} & \textbf{AVA-ActiveSpeaker}  \\
HiGate & MAL & OPP & (mAP) & (mAP) \\
\midrule
\ding{55} & \ding{55} & \ding{55} & 72.49 & 93.95 \\
\ding{55} & \ding{51} & \ding{55} & 73.30 & 93.97 \\
\ding{55} & \ding{55} & \ding{51} & 74.34 & 94.04 \\
\ding{51} & \ding{55} & \ding{55} & 76.33 & 94.54 \\
\ding{51} & \ding{51} & \ding{55} & \underline{76.57} & \underline{94.92} \\
\ding{51} & \ding{51} & \ding{51} & \textbf{77.80} & \textbf{95.05} \\
\bottomrule
\end{tabular}
\caption{Ablation results of HiGate, MAL, and OPP. We report the mean average precision (mAP) for both Ego4D-ASD~\cite{grauman2022ego4d} and AVA-ActiveSpeaker~\cite{roth2020ava}. The baseline uses a simple sum operation for feature fusion followed by two-layer classifier.}
\label{tab:ablation_higate}
% \vspace{-5pt}
\end{table}

\vspace{6pt}\noindent\textbf{Component-wise Ablation.} Table~\ref{tab:ablation_higate} ablates the contribution of each key component in GateFusion on both Ego4D-ASD and AVA-ActiveSpeaker benchmarks.

Starting from the baseline which directly sums the two modality features and passes them to the two-layer classifier, we observe mAP scores of 72.49\% on Ego4D-ASD and 93.95\% on AVA-ActiveSpeaker. Adding either MAL or OPP individually yields incremental improvements, with OPP providing the largest single-component gain, increasing mAP to 74.34\% on Ego4D-ASD and 94.04\% on AVA-ActiveSpeaker. This highlights the benefit of penalizing overconfident visual predictions, especially in egocentric videos.

Introducing HiGate alone results in a substantial performance boost, raising mAP to 76.33\% on Ego4D-ASD and 94.54\% on AVA-ActiveSpeaker, corresponding to improvements of 3.84\% and 0.59\% over the baseline, respectively. Combining HiGate with MAL further enhances performance, reaching 76.57\% on Ego4D-ASD and 94.92\% on AVA-ActiveSpeaker, which demonstrates the value of enforcing unimodal-multimodal consistency. The best results are achieved when all three components are used together, with 77.80\% mAP on Ego4D-ASD and 95.05\% on AVA-ActiveSpeaker benchmarks, reflecting absolute gains of 5.31 and 1.10 points over the baseline.

These results confirm that each component provides complementary benefits, and their combination consistently delivers the best performance. Also, the proposed MAL and OPP improve the model's robustness; more details are provided in Supplementary Sec.\ C. The improvements, particularly on the challenging Ego4D-ASD dataset, underscore the effectiveness and robustness of our proposed model.

\begin{figure}[t]
    \centering
    \includegraphics[width=1\linewidth]{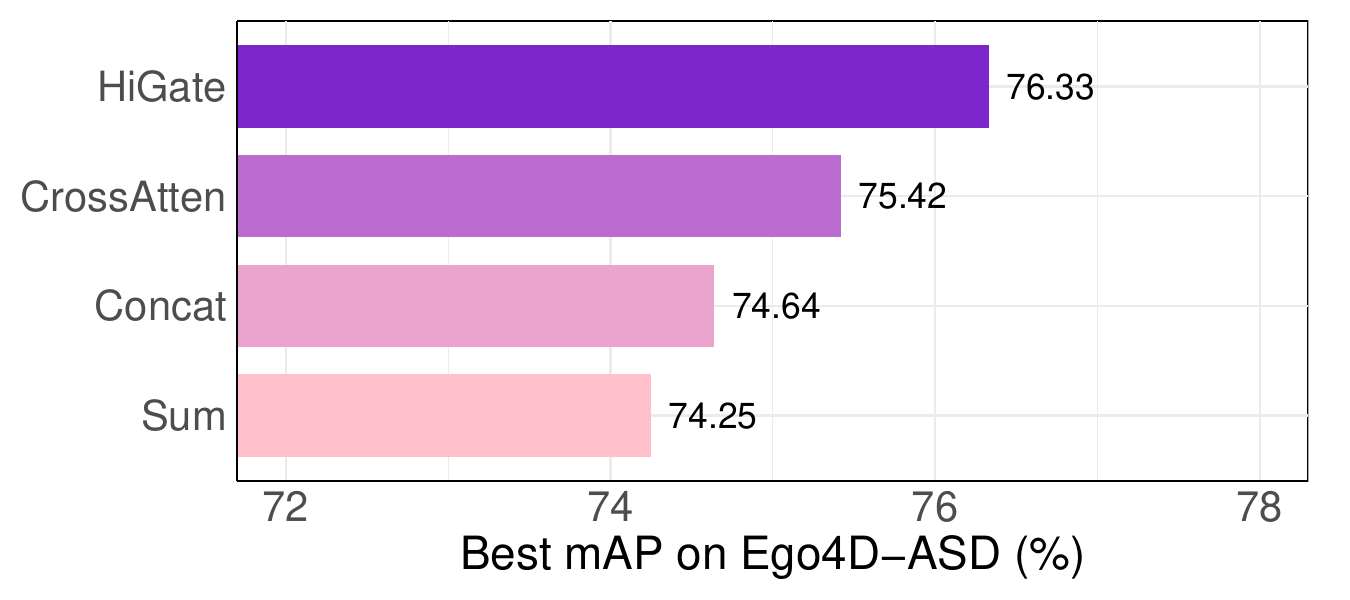}
    \caption{Comparison of mAP scores for different decoders: our HiGate, CrossAtten, Concat, and Sum decoders.}
    \label{fig:decoder}
    % \vspace{-10pt}
\end{figure}

\vspace{6pt}\noindent
\textbf{Comparison with Alternative Decoders.} We further compare our proposed HiGate with representative fusion decoders \textit{without} the auxiliary losses, as illustrated in Fig.~\ref{fig:decoder}. All decoders use the same two-layer classifier where the Concat decoder and Sum decoder have an additional projection layer. The Sum decoder applies simple element-wise summation of multimodal features, while the Concat decoder concatenates features from both modalities along the feature dimension. The CrossAtten decoder implements bidirectional cross-attention, alternating each modality as the query. The resulting features are concatenated and fed into a self-attention layer before being passed through the final classifier.

The results show that the HiGate achieves the highest performance, reaching an mAP of 76.33\% on the Ego4D-ASD benchmark. In comparison, the CrossAtten decoder achieves 75.42\%, while the simple fusion baselines Concat and Sum obtain 74.64\% and 74.25\%, respectively. These results clearly demonstrate that HiGate consistently outperforms both basic and more advanced fusion alternatives, underscoring its effectiveness for cross-modal integration in challenging egocentric scenarios. Supplementary Sec.\ D, E and F provide frozen encoder performance, comparisons with other fusion mechanisms, and efficiency analysis.
\section{Conclusion}
We proposed a new model for active speaker detection. The model includes a novel fusion module, HiGate, that enables efficient, adaptive cross-modal integration at multiple hidden layers and better leverages pretrained encoders. Combined with MAL and OPP, two auxiliary objectives for our multimodal training, our approach achieves robust performance across both egocentric and in-the-wild datasets, surpasses strong baselines and prior fusion strategies, and demonstrates strong generalization in out-of-domain tests. Ablation studies highlight the complementary benefits of each component, especially under challenging conditions. Overall, our results indicate that hierarchical gated fusion is an effective solution for active speaker detection and may hold promise for broader multimodal applications.

\section*{Acknowledgments}
This work was supported in part by the National Science Foundation under award DRL-2112635 to the AI Institute for Engaged Learning. Any opinions, findings, and conclusions or recommendations expressed in this material are those of the author(s) and do not necessarily reflect the views of the National Science Foundation. 

{
    \small
    \bibliographystyle{ieeenat_fullname}
    \bibliography{main}

@String(ICCV= {Int. Conf. Comput. Vis.})

@String(ECCV= {Eur. Conf. Comput. Vis.})

@String(ICASSP=	{ICASSP})

@String(ICCV  = {ICCV})

@String(ECCV  = {ECCV})

@inproceedings{huh2025advancing,
  title={Advancing active speaker detection for egocentric videos},
  author={Huh, Jaesung and Ortiz, Juan Azcarreta and Kumar, Anurag and Pandey, Ashutosh and Aroudi, Ali and Wong, Daniel DE and Nesta, Francesco and Xu, Buye and Donley, Jacob},
  booktitle={ICASSP 2025-2025 IEEE International Conference on Acoustics, Speech and Signal Processing (ICASSP)},
  pages={1--5},
  year={2025},
  organization={IEEE}
}

@article{nguyen2025laser,
  title={LASER: Lip Landmark Assisted Speaker Detection for Robustness},
  author={Nguyen, Le Thien Phuc and Yu, Zhuoran and Lee, Yong Jae},
  journal={arXiv preprint arXiv:2501.11899},
  year={2025}
}

@article{roxo2024asdnb,
  title={ASDnB: Merging Face with Body Cues For Robust Active Speaker Detection},
  author={Roxo, Tiago and Costa, Joana C and In{\'a}cio, Pedro and Proen{\c{c}}a, Hugo},
  journal={arXiv preprint arXiv:2412.08594},
  year={2024}
}

@inproceedings{wang2024loconet,
  title={Loconet: Long-short context network for active speaker detection},
  author={Wang, Xizi and Cheng, Feng and Bertasius, Gedas},
  booktitle={Proceedings of the IEEE/CVF Conference on Computer Vision and Pattern Recognition},
  pages={18462--18472},
  year={2024}
}

@article{min2023sthg,
  title={Sthg: Spatial-temporal heterogeneous graph learning for advanced audio-visual diarization},
  author={Min, Kyle},
  journal={arXiv preprint arXiv:2306.10608},
  year={2023}
}

@inproceedings{zhang2021unicon,
  title={Unicon: Unified context network for robust active speaker detection},
  author={Zhang, Yuanhang and Liang, Susan and Yang, Shuang and Liu, Xiao and Wu, Zhongqin and Shan, Shiguang and Chen, Xilin},
  booktitle={Proceedings of the 29th ACM international conference on multimedia},
  pages={3964--3972},
  year={2021}
}

@inproceedings{cheng2020look,
  title={Look, listen, and attend: Co-attention network for self-supervised audio-visual representation learning},
  author={Cheng, Ying and Wang, Ruize and Pan, Zhihao and Feng, Rui and Zhang, Yuejie},
  booktitle={Proceedings of the 28th ACM International Conference on Multimedia},
  pages={3884--3892},
  year={2020}
}

@inproceedings{awan2024attend,
  title={Attend-Fusion: Efficient Audio-Visual Fusion for Video Classification},
  author={Awan, Mahrukh and Nadeem, Asmar and Awan, Muhammad Junaid and Mustafa, Armin and Husain, Syed Sameed},
  booktitle={European Conference on Computer Vision},
  pages={195--213},
  year={2024},
  organization={Springer}
}

@inproceedings{li2025egonet,
  title={EgoNet: An Unified Egocentric Active Speaker Detection Framework for both Camera Wearer and Visible Candidates},
  author={Li, Yongqian and Zhou, Xin and He, Zheng and Yu, Wei and Luo, Yong},
  booktitle={ICASSP 2025-2025 IEEE International Conference on Acoustics, Speech and Signal Processing (ICASSP)},
  pages={1--5},
  year={2025},
  organization={IEEE}
}

@INPROCEEDINGS{alcazar2021maas,
  author={Alcázar, Juan León and Heilbron, Fabian Caba and Thabet, Ali K. and Ghanem, Bernard},
  booktitle={2021 IEEE/CVF International Conference on Computer Vision (ICCV)}, 
  title={MAAS: Multi-modal Assignation for Active Speaker Detection}, 
  year={2021},
  volume={},
  number={},
  pages={265-274},
  doi={10.1109/ICCV48922.2021.00033}
}

@inproceedings{alcazar2022end,
  title={End-to-end active speaker detection},
  author={Alcazar, Juan Leon and Cordes, Moritz and Zhao, Chen and Ghanem, Bernard},
  booktitle={European Conference on Computer Vision},
  pages={126--143},
  year={2022},
  organization={Springer}
}

@article{shi2022robust,
  title={Robust self-supervised audio-visual speech recognition},
  author={Shi, Bowen and Hsu, Wei-Ning and Mohamed, Abdelrahman},
  journal={arXiv preprint arXiv:2201.01763},
  year={2022}
}

@inproceedings{wuerkaixi2022rethinking,
  title={Rethinking audio-visual synchronization for active speaker detection},
  author={Wuerkaixi, Abudukelimu and Zhang, You and Duan, Zhiyao and Zhang, Changshui},
  booktitle={2022 IEEE 32nd international workshop on machine learning for signal processing (MLSP)},
  pages={01--06},
  year={2022},
  organization={IEEE}
}

@article{vasireddy2024robust,
  title={Robust active speaker detection in noisy environments},
  author={Vasireddy, Siva Sai Nagender and Zhang, Chenxu and Guo, Xiaohu and Tian, Yapeng},
  journal={arXiv preprint arXiv:2403.19002},
  year={2024}
}

@inproceedings{roth2020ava,
  title={Ava active speaker: An audio-visual dataset for active speaker detection},
  author={Roth, Joseph and Chaudhuri, Sourish and Klejch, Ondrej and Marvin, Radhika and Gallagher, Andrew and Kaver, Liat and Ramaswamy, Sharadh and Stopczynski, Arkadiusz and Schmid, Cordelia and Xi, Zhonghua and others},
  booktitle={ICASSP 2020-2020 IEEE International Conference on Acoustics, Speech and Signal Processing (ICASSP)},
  pages={4492--4496},
  year={2020},
  organization={IEEE}
}

@inproceedings{grauman2022ego4d,
  title={Ego4d: Around the world in 3,000 hours of egocentric video},
  author={Grauman, Kristen and Westbury, Andrew and Byrne, Eugene and Chavis, Zachary and Furnari, Antonino and Girdhar, Rohit and Hamburger, Jackson and Jiang, Hao and Liu, Miao and Liu, Xingyu and others},
  booktitle={Proceedings of the IEEE/CVF conference on computer vision and pattern recognition},
  pages={18995--19012},
  year={2022}
}

@article{liao2025lr,
  title={Lr-asd: Lightweight and robust network for active speaker detection},
  author={Liao, Junhua and Duan, Haihan and Feng, Kanghui and Zhao, Wanbing and Yang, Yanbing and Chen, Liangyin and Chen, Yanru},
  journal={International Journal of Computer Vision},
  volume={133},
  number={7},
  pages={4749--4769},
  year={2025},
  publisher={Springer}
}

@article{shi2022learning,
  title={Learning audio-visual speech representation by masked multimodal cluster prediction},
  author={Shi, Bowen and Hsu, Wei-Ning and Lakhotia, Kushal and Mohamed, Abdelrahman},
  journal={arXiv preprint arXiv:2201.02184},
  year={2022}
}

@inproceedings{radford2023robust,
  title={Robust speech recognition via large-scale weak supervision},
  author={Radford, Alec and Kim, Jong Wook and Xu, Tao and Brockman, Greg and McLeavey, Christine and Sutskever, Ilya},
  booktitle={International conference on machine learning},
  pages={28492--28518},
  year={2023},
  organization={PMLR}
}

@inproceedings{tao2021someone,
  title={Is someone speaking? exploring long-term temporal features for audio-visual active speaker detection},
  author={Tao, Ruijie and Pan, Zexu and Das, Rohan Kumar and Qian, Xinyuan and Shou, Mike Zheng and Li, Haizhou},
  booktitle={Proceedings of the 29th ACM international conference on multimedia},
  pages={3927--3935},
  year={2021}
}

@article{loshchilov2017decoupled,
  title={Decoupled weight decay regularization},
  author={Loshchilov, Ilya and Hutter, Frank},
  journal={arXiv preprint arXiv:1711.05101},
  year={2017}
}

@article{min2022intel,
  title={Intel labs at ego4d challenge 2022: A better baseline for audio-visual diarization},
  author={Min, Kyle},
  journal={arXiv preprint arXiv:2210.07764},
  year={2022}
}

@inproceedings{min2022learning,
  title={Learning long-term spatial-temporal graphs for active speaker detection},
  author={Min, Kyle and Roy, Sourya and Tripathi, Subarna and Guha, Tanaya and Majumdar, Somdeb},
  booktitle={European conference on computer vision},
  pages={371--387},
  year={2022},
  organization={Springer}
}

@inproceedings{liao2023light,
  title={A light weight model for active speaker detection},
  author={Liao, Junhua and Duan, Haihan and Feng, Kanghui and Zhao, Wanbing and Yang, Yanbing and Chen, Liangyin},
  booktitle={Proceedings of the IEEE/CVF conference on computer vision and pattern recognition},
  pages={22932--22941},
  year={2023}
}

@inproceedings{kopuklu2021design,
  title={How to design a three-stage architecture for audio-visual active speaker detection in the wild},
  author={K{\"o}p{\"u}kl{\"u}, Okan and Taseska, Maja and Rigoll, Gerhard},
  booktitle={Proceedings of the IEEE/CVF international conference on computer vision},
  pages={1193--1203},
  year={2021}
}

@inproceedings{kim2021look,
  title={Look who's talking: Active speaker detection in the wild},
  author={Kim, You Jin and Heo, Hee Soo and Choe, Soyeon and Chung, Soo Whan and Kwon, Yoohwan and Lee, Bong Jin and Kwon, Youngki and Chung, Joon Son},
  booktitle={22nd Annual Conference of the International Speech Communication Association, INTERSPEECH 2021},
  pages={4411--4415},
  year={2021},
  organization={International Speech Communication Association}
}

@inproceedings{jung2024talknce,
  title={Talknce: Improving active speaker detection with talk-aware contrastive learning},
  author={Jung, Chaeyoung and Lee, Suyeon and Nam, Kihyun and Rho, Kyeongha and Kim, You Jin and Jang, Youngjoon and Chung, Joon Son},
  booktitle={ICASSP 2024-2024 IEEE International Conference on Acoustics, Speech and Signal Processing (ICASSP)},
  pages={8391--8395},
  year={2024},
  organization={IEEE}
}

@inproceedings{jeong2025learning,
  title={Learning Audio-guided Video Representation with Gated Attention for Video-Text Retrieval},
  author={Jeong, Boseung and Park, Jicheol and Kim, Sungyeon and Kwak, Suha},
  booktitle={Proceedings of the Computer Vision and Pattern Recognition Conference},
  pages={26202--26211},
  year={2025}
}

@inproceedings{alcazar2020active,
  title={Active speakers in context},
  author={Alc{\'a}zar, Juan Le{\'o}n and Caba, Fabian and Mai, Long and Perazzi, Federico and Lee, Joon-Young and Arbel{\'a}ez, Pablo and Ghanem, Bernard},
  booktitle={Proceedings of the IEEE/CVF conference on computer vision and pattern recognition},
  pages={12465--12474},
  year={2020}
}

@inproceedings{lin2023av,
  title={Av-sepformer: Cross-attention sepformer for audio-visual target speaker extraction},
  author={Lin, Jiuxin and Cai, Xinyu and Dinkel, Heinrich and Chen, Jun and Yan, Zhiyong and Wang, Yongqing and Zhang, Junbo and Wu, Zhiyong and Wang, Yujun and Meng, Helen},
  booktitle={ICASSP 2023-2023 IEEE International Conference on Acoustics, Speech and Signal Processing (ICASSP)},
  pages={1--5},
  year={2023},
  organization={IEEE}
}

@inproceedings{wang2024mlca,
  title={Mlca-avsr: Multi-layer cross attention fusion based audio-visual speech recognition},
  author={Wang, He and Guo, Pengcheng and Zhou, Pan and Xie, Lei},
  booktitle={ICASSP 2024-2024 IEEE International Conference on Acoustics, Speech and Signal Processing (ICASSP)},
  pages={8150--8154},
  year={2024},
  organization={IEEE}
}

@article{wang2024catnet,
  title={CATNet: Cross-modal fusion for audio--visual speech recognition},
  author={Wang, Xingmei and Mi, Jiachen and Li, Boquan and Zhao, Yixu and Meng, Jiaxiang},
  journal={Pattern Recognition Letters},
  volume={178},
  pages={216--222},
  year={2024},
  publisher={Elsevier}
}

@inproceedings{xu2024rethink,
  title={Rethink cross-modal fusion in weakly-supervised audio-visual video parsing},
  author={Xu, Yating and Hu, Conghui and Lee, Gim Hee},
  booktitle={Proceedings of the IEEE/CVF Winter Conference on Applications of Computer Vision},
  pages={5615--5624},
  year={2024}
}

@inproceedings{wortwein2024smurf,
  title={Smurf: Statistical modality uniqueness and redundancy factorization},
  author={W{\"o}rtwein, Torsten and Allen, Nicholas B and Cohn, Jeffrey F and Morency, Louis-Philippe},
  booktitle={Proceedings of the 26th International Conference on Multimodal Interaction},
  pages={339--349},
  year={2024}
}

@inproceedings{datta2022asd,
  title={Asd-transformer: Efficient active speaker detection using self and multimodal transformers},
  author={Datta, Gourav and Etchart, Tyler and Yadav, Vivek and Hedau, Varsha and Natarajan, Pradeep and Chang, Shih-Fu},
  booktitle={ICASSP 2022-2022 IEEE International Conference on Acoustics, Speech and Signal Processing (ICASSP)},
  pages={4568--4572},
  year={2022},
  organization={IEEE}
}

@article{dataset_wasd,
  title={WASD: A Wilder Active Speaker Detection Dataset},
  author={Roxo, Tiago and Costa, Joana C and In{\'a}cio, Pedro RM and Proen{\c{c}}a, Hugo},
  journal={IEEE Transactions on Biometrics, Behavior, and Identity Science},
  year={2024},
  publisher={IEEE},
  doi={10.1109/TBIOM.2024.3412821}
}

@article{dataset_unitalk,
  title={UniTalk: Towards Universal Active Speaker Detection in Real World Scenarios},
  author={Nguyen, Le Thien Phuc and Yu, Zhuoran and Cao, Khoa Quang Nhat and Guo, Yuwei and Pham, Tu Ho Manh and Nguyen, Tuan Tai and Vo, Toan Ngo Duc and Poon, Lucas and Lee, Soochahn and Lee, Yong Jae},
  journal={arXiv preprint arXiv:2505.21954},
  year={2025}
}

@inproceedings{ts_talknet,
  title={Target Active Speaker Detection with Audio-visual Cues},
  author={Jiang, Yidi and Tao, Ruijie and Pan, Zexu and Li, Haizhou},
  booktitle={Proc. Interspeech},
  year={2023}
}

@article{method_bias,
  title={BIAS: A Body-Based Interpretable Active Speaker Approach}, 
  author={Roxo, Tiago and Costa, Joana C. and Inácio, Pedro R. M. and Proença, Hugo},
  journal={IEEE Transactions on Biometrics, Behavior, and Identity Science}, 
  year={2024},
  publisher={IEEE},
  doi={10.1109/TBIOM.2024.3520030}
}

@article{xiong2022looklistenmultimodalcorrelationlearning,
  title={Look\&listen: Multi-modal correlation learning for active speaker detection and speech enhancement},
  author={Xiong, Junwen and Zhou, Yu and Zhang, Peng and Xie, Lei and Huang, Wei and Zha, Yufei},
  journal={IEEE Transactions on Multimedia},
  volume={25},
  pages={5800--5812},
  year={2022},
  publisher={IEEE}
}

@article{lin2023quavf,
  title={Quavf: Quality-aware audio-visual fusion for ego4d talking to me challenge},
  author={Lin, Hsi-Che and Wang, Chien-Yi and Chen, Min-Hung and Fu, Szu-Wei and Wang, Yu-Chiang Frank},
  journal={arXiv preprint arXiv:2306.17404},
  year={2023}
}

@inproceedings{owens2018audio,
  title={Audio-visual scene analysis with self-supervised multisensory features},
  author={Owens, Andrew and Efros, Alexei A},
  booktitle={Proceedings of the European conference on computer vision (ECCV)},
  pages={631--648},
  year={2018}
}

@inproceedings{mo2024unveiling,
  title={Unveiling the power of audio-visual early fusion transformers with dense interactions through masked modeling},
  author={Mo, Shentong and Morgado, Pedro},
  booktitle={Proceedings of the IEEE/CVF Conference on Computer Vision and Pattern Recognition},
  pages={27186--27196},
  year={2024}
}

@inproceedings{he2016deep,
  title={Deep residual learning for image recognition},
  author={He, Kaiming and Zhang, Xiangyu and Ren, Shaoqing and Sun, Jian},
  booktitle={Proceedings of the IEEE conference on computer vision and pattern recognition},
  pages={770--778},
  year={2016}
}

@inproceedings{tran2018closer,
  title={A closer look at spatiotemporal convolutions for action recognition},
  author={Tran, Du and Wang, Heng and Torresani, Lorenzo and Ray, Jamie and LeCun, Yann and Paluri, Manohar},
  booktitle={Proceedings of the IEEE conference on Computer Vision and Pattern Recognition},
  pages={6450--6459},
  year={2018}
}

@inproceedings{hershey2017cnn,
  title={CNN architectures for large-scale audio classification},
  author={Hershey, Shawn and Chaudhuri, Sourish and Ellis, Daniel PW and Gemmeke, Jort F and Jansen, Aren and Moore, R Channing and Plakal, Manoj and Platt, Devin and Saurous, Rif A and Seybold, Bryan and others},
  booktitle={2017 ieee international conference on acoustics, speech and signal processing (icassp)},
  pages={131--135},
  year={2017},
  organization={IEEE}
}

@inproceedings{radford2021learning,
  title={Learning transferable visual models from natural language supervision},
  author={Radford, Alec and Kim, Jong Wook and Hallacy, Chris and Ramesh, Aditya and Goh, Gabriel and Agarwal, Sandhini and Sastry, Girish and Askell, Amanda and Mishkin, Pamela and Clark, Jack and others},
  booktitle={International conference on machine learning},
  pages={8748--8763},
  year={2021},
  organization={PMLR}
}

@inproceedings{gong21b_interspeech,
  author={Yuan Gong and Yu-An Chung and James Glass},
  title={{AST: Audio Spectrogram Transformer}},
  year=2021,
  booktitle={Proc. Interspeech 2021},
  pages={571--575},
  doi={10.21437/Interspeech.2021-698}
}

@inproceedings{deng2009imagenet,
  title={Imagenet: A large-scale hierarchical image database},
  author={Deng, Jia and Dong, Wei and Socher, Richard and Li, Li-Jia and Li, Kai and Fei-Fei, Li},
  booktitle={2009 IEEE conference on computer vision and pattern recognition},
  pages={248--255},
  year={2009},
  organization={Ieee}
}

@inproceedings{kopuklu2019resource,
  title={Resource efficient 3d convolutional neural networks},
  author={Kopuklu, Okan and Kose, Neslihan and Gunduz, Ahmet and Rigoll, Gerhard},
  booktitle={Proceedings of the IEEE/CVF international conference on computer vision workshops},
  pages={0--0},
  year={2019}
}

@inproceedings{hara2018can,
  title={Can spatiotemporal 3d cnns retrace the history of 2d cnns and imagenet?},
  author={Hara, Kensho and Kataoka, Hirokatsu and Satoh, Yutaka},
  booktitle={Proceedings of the IEEE conference on Computer Vision and Pattern Recognition},
  pages={6546--6555},
  year={2018}
}

@inproceedings{desplanques2020ecapa,
  title={ECAPA-TDNN: Emphasized Channel Attention, Propagation and Aggregation in TDNN Based Speaker Verification},
  author={Desplanques, Brecht and Thienpondt, Jenthe and Demuynck, Kris},
  booktitle={Proc. Interspeech 2020},
  pages={3830--3834},
  year={2020}
}

@article{qiu2025gated,
  title={Gated Attention for Large Language Models: Non-linearity, Sparsity, and Attention-Sink-Free},
  author={Qiu, Zihan and Wang, Zekun and Zheng, Bo and Huang, Zeyu and Wen, Kaiyue and Yang, Songlin and Men, Rui and Yu, Le and Huang, Fei and Huang, Suozhi and others},
  journal={arXiv preprint arXiv:2505.06708},
  year={2025}
}
}

\setcounter{section}{0}
\renewcommand{\thesection}{\Alph{section}}
\twocolumn[
\begin{center}
    % Title
    \textbf{\Large Supplementary Material for GateFusion: Hierarchical Gated Cross-Modal Fusion for Active Speaker Detection} \\
    \vspace{1.5em}

%     % Author Names
%     \large
%     Yu Wang \quad Juhyung Ha \quad Frangil M. Ramirez \quad Yuchen Wang \quad David J. Crandall \\
%     \vspace{0.5em}
    
%     % Affiliation & Email
%     \normalsize
%     Indiana University \\
%     Bloomington, Indiana, USA \\
%     {\tt\small \{yw173, juhha, fraramir, wang617, djcran\}@iu.edu}
\end{center}
\vspace{20pt}
]

% \thispagestyle{empty}
% \vspace{-50pt}
\section{Pretrained Backbone Configurations}

\begin{table}[h]
    \centering
    \footnotesize
    \setlength{\tabcolsep}{3pt}
    \begin{tabularx}{\columnwidth}{X c c}
    \toprule
    \textbf{Method} & \textbf{Pretrain-V} & \textbf{Pretrain-A} \\
    \midrule
    ASC (CVPR'20)~\cite{alcazar2020active} & ResNet-18~\cite{he2016deep} & N/A \\
    TalkNet (MM'21)~\cite{tao2021someone} & N/A & N/A \\
    ASDNet (ICCV'21)~\cite{kopuklu2021design} & 3D-ResNeXt~\cite{kopuklu2019resource} & N/A \\
    MAAS (ICCV'21)~\cite{alcazar2021maas} & ResNet-18 & ResNet-18 \\
    EASEE-50 (ECCV'22)~\cite{alcazar2022end} & 3D-ResNet~\cite{hara2018can} & ResNet-18 \\
    SPELL (ECCV'22)~\cite{min2022intel, min2022learning} & N/A & N/A \\
    Sync-TalkNet (MLSP'22)~\cite{wuerkaixi2022rethinking} & N/A & ResNet-34~\cite{he2016deep} \\
    ASD-Transformer (ICASSP'22)~\cite{datta2022asd} & N/A & N/A \\
    LightASD (CVPR'23)~\cite{liao2023light} & N/A & N/A \\
    STHG (CVPRW'23)~\cite{min2023sthg} & N/A & N/A \\
    TS-TalkNet (Interspeech'23)~\cite{ts_talknet} & N/A & N/A \\
    TalkNCE (ICASSP'24)~\cite{jung2024talknce} & N/A & VGGish~\cite{hershey2017cnn} \\
    BIAS (T-BIOM'24)~\cite{method_bias} & N/A & N/A \\
    ASDnB (arXiv'24)~\cite{roxo2024asdnb} & N/A & N/A \\
    LoCoNet (CVPR'24)~\cite{wang2024loconet} & N/A & VGGish \\
    LR-ASD (IJCV'25)~\cite{liao2025lr} & N/A & N/A \\
    \bottomrule
    \end{tabularx}
    \caption{
    Pretrained backbone configurations of state-of-the-art ASD models. \textbf{Pretrain-V/A} denote pretrained weights for video/audio encoders (N/A: trained from scratch). For TalkNCE, we use its strongest LoCoNet-based variant. TS-TalkNet trains encoders from scratch but includes a pretrained ECAPA-TDNN~\cite{desplanques2020ecapa} for extracting target-speaker embeddings. Though initialized from scratch, ASDnB is pretrained on WASD~\cite{dataset_wasd} for AVA-ActiveSpeaker~\cite{roth2020ava}, and LoCoNet on AVA-ActiveSpeaker for Ego4D-ASD~\cite{grauman2022ego4d}.}
    \label{tab:backbones}

    \vspace{10pt} 
    % \small
    \centering
    \begin{tabularx}{\columnwidth}{X c c}
        \toprule
        \textbf{Method} & \textbf{mAP (Baseline)} & \textbf{mAP (+ Pretrained Enc.)} \\
        \midrule
        LoCoNet~\cite{wang2024loconet} & 59.3 & 60.7 \\
        TalkNet~\cite{tao2021someone} & 51.7 & 70.7 \\
        \bottomrule
    \end{tabularx}
    \caption{Performance of representative ASD models before and after adopting our pretrained video and audio encoders on the Ego4D-ASD benchmark. Note that LoCoNet is trained from scratch here.}
    \label{tab:fairness}
% \vspace{-10pt}
\end{table}

We summarize in Table~\ref{tab:backbones} the pretrained backbone configurations adopted by the state-of-the-art models. Prior works exhibit variability in their initialization strategies: some rely on pretrained encoders for their visual or audio branches, e.g., a ResNet-18~\cite{he2016deep} pretrained on ImageNet~\cite{deng2009imagenet}, while others train all components strictly from scratch. Our model adopts pretrained weights for both video and audio encoders (we retain only the first 12 layers of each pretrained checkpoint to match the depth of our encoders). To verify that our performance improvements do not arise solely from pretrained initialization, we additionally equip two representative baselines, TalkNet~\cite{tao2021someone} and LoCoNet~\cite{wang2024loconet}, with the same pretrained video and audio encoders used in our framework and evaluate all models under identical conditions on the Ego4D-ASD benchmark~\cite{grauman2022ego4d}. This dataset is particularly challenging and serves as a rigorous testbed for assessing the true utility of pretrained backbones. As reported in Table~\ref{tab:fairness}, simply adding our pretrained encoders to existing architectures improves performance but does not match the gains achieved by our full model, thereby demonstrating that our contributions extend beyond encoder initialization.

\section{Additional Ablation Studies}
\begin{figure}[ht]
  \centering
  \includegraphics[width=1\linewidth]{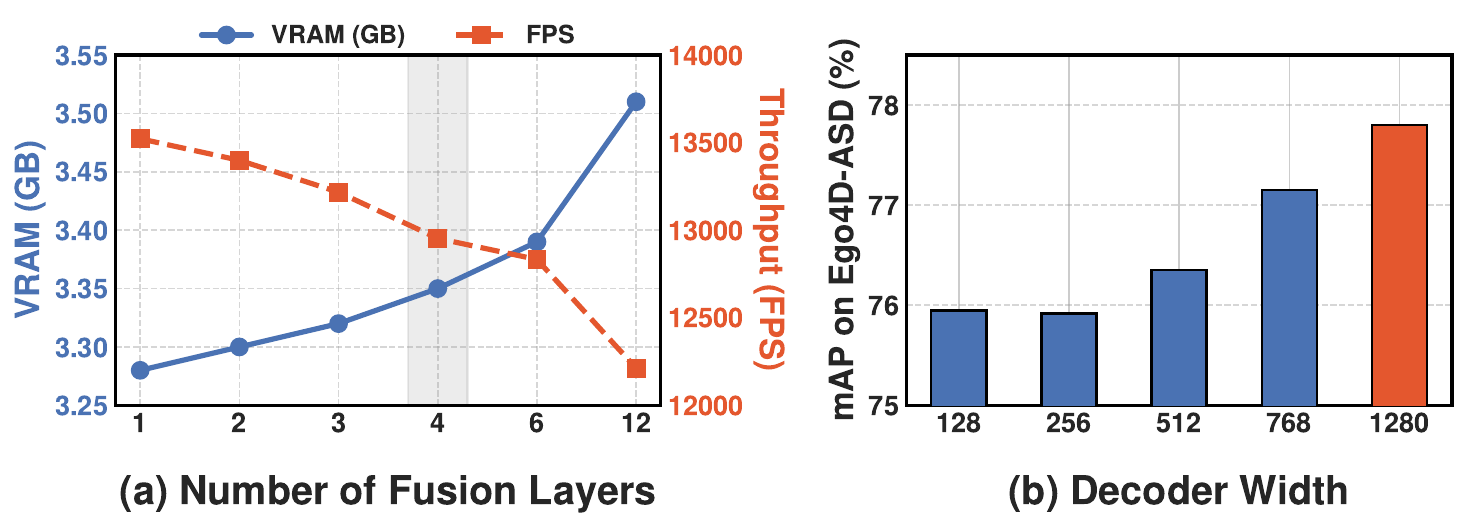}
  \caption{Ablations on model hyperparameters. (a) We visualize the trade-off between memory cost (VRAM, blue solid line) and inference throughput (FPS, orange dashed line) as the number of fusion layers increases. The gray band highlights our selected configuration ($N=4$). (b) Performance across different decoder widths.}
  \label{fig:ablate_handpick}
  % \vspace{-10pt}
\end{figure}

\begin{table}[t]
\centering
\vspace{-5pt}
\small
\begin{tabularx}{\columnwidth}{X r}
\toprule
\textbf{\#Fusion Layers} & \textbf{Layer Index} \\
\midrule
1  & [10] \\
2  & [7, 10] \\
3  & [4, 7, 10] \\
4  & [1, 4, 7, 10] \\
6  & [1, 3, 5, 7, 9, 11] \\
12 & [1, 2, 3, 4, 5, 6, 7, 8, 9, 10, 11, 12] \\
\bottomrule
\end{tabularx}
\caption{Fusion layer configurations for the ablation on the number of fusion layers. \textbf{Layer Index} lists the indices (starting from 1) of the Transformer blocks used for fusion.}
\label{tab:numberlayer}
\end{table}

We provide additional ablations regarding the number of fusion layers and decoder width on the Ego4D-ASD benchmark~\cite{grauman2022ego4d}, as illustrated in Fig.~\ref{fig:ablate_handpick}. The models are trained with the MAL and OPP objectives.

As discussed in the main paper, increasing $N$ generally improves model capacity. The specific configurations are detailed in Table~\ref{tab:numberlayer}, ranging from single-layer fusion to fully dense fusion ($N=12$). However, Fig.~\ref{fig:ablate_handpick}(a) reveals the associated computational costs: a larger $N$ leads to increased memory footprint and a corresponding drop in inference throughput. We select $N=4$ as it strikes a balance between efficiency and performance. Fig.~\ref{fig:ablate_handpick}(b) shows that increasing decoder width steadily improves performance. We adopt 1280 for optimal results.

\section{Additional Losses for Model Robustness}

\begin{table}[ht]
\centering
\small
\setlength{\tabcolsep}{6pt}
\begin{tabularx}{\linewidth}{X|ccc}
\toprule
\multirow{2}{*}{\textbf{Method}} & \multicolumn{3}{c}{\textbf{AV Prediction (mAP)}} \\ \cmidrule(lr){2-4}
 & \textbf{Audio Noise} & \textbf{Video Blur} & \textbf{Clean} \\
\midrule
Baseline & 60.10 & 70.80 & 76.33 \\
Baseline+MAL+OPP & \textbf{64.06} & \textbf{71.73} & \textbf{77.80} \\
\bottomrule
\end{tabularx}
\caption{Robustness against data corruption. Comparison of AV prediction performance under audio noise and video blur conditions. Our method demonstrates superior robustness compared to the baseline.} 
\label{tab:noise_robustness}
\end{table}

In this section, we evaluate the robustness of our proposed method against data quality degradation. We simulate two types of degradation: noisy audio and blurry video, reflecting common real-world scenarios. Specifically, we introduce additive white noise to the audio stream. For the video stream, we implement a heterogeneous blur protocol, where Gaussian blur and Radial blur are applied to the dataset in an equal 50/50 split. We compare our full model (incorporating MAL and OPP) against the baseline without these auxiliary losses. 

As shown in Table \ref{tab:noise_robustness}, our proposed framework demonstrates superior stability under sensory corruption. In the Audio Noise setting, the baseline suffers a relative performance drop of 21.3\% compared to the clean setting, whereas our full model limits the degradation to 17.7\%. Furthermore, under the Video Blur condition, our method consistently maintains a performance advantage over the baseline (71.73\% vs. 70.80\%). These results indicate that the proposed auxiliary constraints effectively enable the model to recover cues from the corrupted modality by leveraging the context of the clean modality.

We further evaluate the extreme case of complete modality loss, observing a marginal degradation compared to the baseline. This expected trade-off evidences the deeper multimodal interaction learned by our method. Unlike the baseline which processes modalities independently, our approach enforces tight coupling. Consequently, the fusion module relies on learned cross-modal synergy, rendering a missing modality an out-of-distribution input rather than a simple information loss.

\section{Frozen Encoder Performance}
To assess the impact of encoder fine-tuning, we conduct an experiment where both the video and audio encoders are entirely frozen and only the decoder parameters are trained. In this setting, the model achieves an mAP of 68.16\% on the Ego4D-ASD benchmark. Although this is lower than our full fine-tuning setup (76.33\% mAP), the performance remains competitive compared to prior works such as LoCoNet (68.4\%) and SPELL (60.7\%), despite using fixed encoders and no end-to-end optimization.

\section{Comparing with Other Fusion Methods}
\begin{table}[ht]
\centering
\vspace{-5pt}
\small
\begin{tabularx}{\columnwidth}{X c c c}
\toprule
\textbf{Fusion Method} & \textbf{Fusion Stage} & \textbf{\#Layers} & \textbf{mAP (\%)} \\
\midrule
Owens'Fusion~\cite{owens2018audio} & Early & 1 & 37.63 \\
Messenger~\cite{xu2024rethink} & Mid & 1 & 73.81 \\
DeepAVFusion~\cite{mo2024unveiling} & Multi-layer & 12 & 66.06 \\
CATNet~\cite{wang2024catnet} & Multi-layer & 3 & 68.15 \\
MLCA~\cite{wang2024mlca} & Multi-layer & 3 & 72.01 \\
BiAVIGATE~\cite{jeong2025learning} & Multi-layer & 4 & 73.97 \\
\midrule
HiGate & Multi-layer & 4 & \textbf{76.33} \\
\bottomrule
\end{tabularx}
% \vspace{-10pt}
\caption{Comparison with other audio-visual fusion strategies. BiAVIGATE denotes bidirectional AVIGATE~\cite{jeong2025learning}.}
\label{tab:avfusion}
\end{table}
Early, mid, and multi-layer fusion strategies are rarely explored in ASD but are well studied in related multimodal works. We implement six representative audio-visual fusion methods on our encoders (12-layer truncated AV-HuBERT~\cite{shi2022learning} and Whisper~\cite{radford2023robust}) without task-specific components or additional losses, adhering closely to the original settings and using official code when available. The compared methods include Owens and Efros’ method (Owens’Fusion for simplicity)~\cite{owens2018audio}, Messenger~\cite{xu2024rethink}, DeepAVFusion~\cite{mo2024unveiling}, CATNet~\cite{wang2024catnet}, MLCA~\cite{wang2024mlca}, and bidirectional AVIGATE (BiAVIGATE for simplicity)~\cite{jeong2025learning}. Results are presented in Table~\ref{tab:avfusion}.

We implement each fusion method following the configuration described in the original papers. Owens’Fusion~\cite{owens2018audio} combines shallow unimodal features into a single branch. Accordingly, we remove Whisper~\cite{radford2023robust} Transformer blocks, expand the convolutional front end to extract shallow audio features, and process the fused features with the 12-layer AV-HuBERT~\cite{shi2022learning} branch, which is intrinsically designed for audio-visual input. The absence of a strong audio encoder substantially reduces mAP. 

Both Messenger~\cite{xu2024rethink} and AVIGATE~\cite{jeong2025learning} rely on strong pretrained audio and visual encoders. Messenger uses a pretrained ResNet-152~\cite{he2016deep} and an 18-layer R(2+1)D~\cite{tran2018closer} model for visual features, together with a pretrained VGGish network~\cite{hershey2017cnn} for audio features. AVIGATE employs a pretrained CLIP~\cite{radford2021learning} visual encoder (ViT-B) and a pretrained 12-layer AST~\cite{gong21b_interspeech} for audio. Since both methods already rely on powerful encoders, we replace them with our own and re-implement their fusion blocks on top. In addition, because AVIGATE~\cite{jeong2025learning} treats audio and video asymmetrically, we extend it to bidirectional fusion to align with our framework.

DeepAVFusion~\cite{mo2024unveiling} introduces a fusion stream that integrates multimodal features at each Transformer block of both modalities. We follow this setting to adapt their fusion strategy to our 12-layer encoders. CATNet~\cite{wang2024catnet} and MLCA~\cite{wang2024mlca} both perform multi-stage fusion, but at different depths. For CATNet~\cite{wang2024catnet}, we split our 12-layer encoders into two stages: the first six layers apply shallow fusion and the remaining six apply middle fusion, with the outputs combined through late fusion. For MLCA~\cite{wang2024mlca}, we insert fusion blocks after layers 4, 8, and the final layer of our encoders, following the progressive multi-layer fusion design. Each block applies self-attention within each modality, followed by bidirectional cross-modal attention with residual feedback, before returning the updated features to the encoders.

Table~\ref{tab:avfusion} shows that the early-fusion baseline (Owens’Fusion~\cite{owens2018audio}) performs worst (37.63\% mAP). Dense early fusion across all layers (DeepAVFusion~\cite{mo2024unveiling}, 12 layers) lags at 66.06\% mAP, suggesting that aggressive early coupling is suboptimal for ASD. A single mid-layer fusion (Messenger~\cite{xu2024rethink}) improves performance (73.81\% mAP) but fails to capture stage-specific cues. Multi-layer designs generally help: MLCA~\cite{wang2024mlca} reaches 72.01\% mAP, CATNet~\cite{wang2024catnet} 68.15\% mAP, and our bidirectional adaptation of AVIGATE~\cite{jeong2025learning} attains 73.97\% mAP.  Our proposed HiGate achieves the best performance (76.33\% mAP), outperforming the next best baseline (BiAVIGATE~\cite{jeong2025learning}) by +2.36\% mAP, and surpassing Messenger~\cite{xu2024rethink} and MLCA~\cite{wang2024mlca} by +2.52\% and +4.32\% mAP, respectively.

\section{Efficiency Analysis}
\begin{table}[ht]
    \centering
    \footnotesize
    \begin{tabularx}{\columnwidth}{Xcccc}
        \toprule
        \multirow[c]{1}{1.6cm}{\textbf{Method}} & \textbf{Video frames} & \textbf{VRAM (GB)} & \textbf{Time (ms)} & \textbf{FPS} \\
        \midrule
        \multirow[c]{4}{1.6cm}{LR-ASD~\cite{liao2025lr}} & 1000 & 1.25 & 38.63 & 25890 \\
        & 2000 & 2.50 & 78.64 & 25431 \\
        & 4000 & 4.99 & 156.09 & 25627 \\
        & 6000 & 7.48 & 236.32 & 25390 \\
        \hline
        \multirow[c]{4}{1.6cm}
        {TalkNet~\cite{kim2021look}} & 1000 & 1.89 & 48.65 & 20556 \\
        & 2000 & 3.52 & 102.52 & 19508 \\
        & 4000 & 6.96 & 213.88 & 18702 \\
        & 6000 & 10.32 & 328.01 & 18292 \\
        \hline
        \multirow[c]{3}{1.6cm}{LoCoNet~\cite{wang2024loconet}} & 1000 & 5.02 & 135.45 & 7383 \\
        & 2000 & 10.19 & 274.97 & 7273 \\
        & 4000 & \multicolumn{3}{c}{Out of Memory} \\
        \hline
        \multirow[c]{4}{1.6cm}{GateFusion} & 1000 & 3.35 & 77.20 & 12953\\
        & 2000 & 4.96 & 174.37 & 11470 \\
        & 4000 & 8.17 & 423.37 & 9448 \\
        & 6000 & 11.37 & 749.56 & 8005 \\
        \bottomrule
    \end{tabularx}
    \caption{Efficiency analysis using inference VRAM consumption, runtime, and FPS. Comparing with previous SOTA, LoCoNet, our model shows better efficiency in all metrics.}
     % \vspace{-15pt}
     \label{tab:efficiency}
\end{table}

We report inference runtime and VRAM consumption in Table~\ref{tab:efficiency} using a single NVIDIA L40S GPU. For this, we followed the evaluation protocol reported by LR-ASD~\cite{liao2025lr}. Ordered from most to least efficient, the models rank as follows: LR-ASD, TalkNet~\cite{tao2021someone}, our proposed GateFusion, and LoCoNet~\cite{wang2024loconet}. Compared to the previous SOTA (LocoNet), GateFusion not only surpasses its performance but also shows greater efficiency in both VRAM usage and runtime. Given 1,000 frames, GateFusion consumes 33\% less VRAM with a 75\% faster runtime.

\end{document}